\documentclass{llncs} 
\usepackage{url}
\usepackage{glossaries}
\glsdisablehyper
\newacronym{nlp}{NLP}{Natural Language Processing}
\newacronym{rnn}{RNN}{Recurrent Neural Network}
\newacronym{lstm}{LSTM}{Long Short-Term Memory network}
\newacronym{gru}{GRU}{Gated Recurrent Unit}
\newacronym{mrnn}{mRNN}{ multiplicative \gls{rnn}}
\newacronym{mlstm}{mLSTM}{ multiplicative \gls{lstm}}
\newacronym{mgru}{mGRU}{multiplicative \gls{gru}}
\newacronym{trnn}{tRNN}{tensor \gls{rnn}}
\newacronym{tmlstm}{tmLSTM}{true \gls{mlstm}}
\newacronym{tmgru}{tmGRU}{true multiplicative \gls{gru}}
\newacronym{mirnn}{MI-RNN}{Multiplicative Integration \gls{rnn}}
\newacronym{milstm}{MI-LSTM}{Multiplicative Integration \gls{lstm}}
\newacronym{rlm}{RLM}{Recurrent Language Model}
\newacronym{bpc}{BPC}{bits per character}

\usepackage[utf8]{inputenc}

\usepackage{caption} 
\usepackage{todonotes}

\begin{document}
\title{Multiplicative Models for Recurrent\\Language Modeling}
\author{Diego Maupomé, Marie-Jean Meurs\\
{\texttt{\small maupome.diego@courrier.uqam.ca\\
meurs.marie-jean@uqam.ca}}\\
}
\institute{Universit\'e du Qu\'ebec \`a Montr\'eal, Montr\'eal, QC, Canada}

\maketitle
\begin{abstract}
Recently, there has been interest in multiplicative recurrent neural networks for language modeling. 
Indeed, simple Recurrent Neural Networks (RNNs) encounter difficulties recovering from past mistakes when generating sequences due to high correlation between hidden states. 
These challenges can be mitigated by integrating second-order terms in the hidden-state update. 
One such model, multiplicative Long Short-Term Memory (mLSTM) is particularly interesting in its original formulation because of the sharing of its second-order term, referred to as the intermediate state.
We explore these architectural improvements by introducing new models and testing them on character-level language modeling tasks.
This allows us to establish the relevance of shared parametrization in recurrent language modeling.
\end{abstract}

\section{Introduction}

One of the principal challenges in computational linguistics is to account for the word order of the document or utterance being processed~\cite{ghodsi2016analysis}.
Of course, the numbers of possible phrases grows exponentially with respect to a given phrase length, requiring an approximate approach to summarizing its content.
\glspl{rnn} are such an approach, and they are used in various tasks in \gls{nlp}, such as machine translation~\cite{luong2015effective}, abstractive summarization~\cite{abssum} and question answering~\cite{qna}.
However, \glspl{rnn}, as approximations, suffer from numerical troubles that have been identified, such as that of recovering from past errors when generating phrases.
We take interest in a model that mitigates this problem, \glspl{mrnn}, and how it has been and can be combined for new models.
To evaluate these models, we use the task of \emph{recurrent language modeling}, which consists in predicting the next token (character or word) in a document.
This paper is organized as follows: \glspl{rnn} and \glspl{mrnn} are introduced respectively in Sections~\ref{sec:rnn} and \ref{sec:mrnn}.
Section~\ref{sec:mmod} presents new and existing multiplicative models.
Section~\ref{sec:exp} describes the datasets and experiments performed, as well as results obtained.
Sections ~\ref{sec:conc} discusses and concludes our findings.

\section{Recurrent neural networks}
\label{sec:rnn}
\glspl{rnn} are powerful tools of sequence modeling that can preserve the order of words or characters in a document.
A document is therefore a sequence of words, $x_1,\ldots,x_T$. 
Given the exponential growth of possible histories with respect to the sequence length, the probability of observing a given sequence needs to be approximated. \glspl{rnn} will make this approximation using the product rule,
\begin{equation*}
    P(x_1,\ldots,x_T) = P(x_1)P(x_2|x_1)\ldots P(x_T|x_1,\ldots, x_{T-1}),
\end{equation*}
and updating a \emph{hidden state} at every time step. 
This state is first null,
\begin{equation*}
    h_0 = \mathbf{0}.
\end{equation*}
Thereafter, it is computed as a function of the past hidden state as well as the input at the current time step,
\begin{equation*}
    h_t = f(h_{t-1}, x_t),
\end{equation*}
known as the \emph{transition function}.
$f$ is a learned function, often taking the form
\begin{equation*}
    h_t = \textrm{tanh}(U x_t +  W h_{t-1}).\footnote{Additive biases are omitted throughout the paper for concision}
\end{equation*}
This allows, in theory, for straightforward modeling of sequences of arbitrary length.

In practice, \glspl{rnn} encounter some difficulties that need some clever engineering to be mitigated. 
For example, learning long-term dependencies such as those found in language is not without its share of woes arising from numerical considerations, such as the well-known vanishing gradient problem~\cite{bengio1994learning}. This can be addressed with gating mechanisms, such as \gls{lstm}~\cite{hochreiter1997long}  and \gls{gru}~\cite{cho2014learning}. 

A problem that is more specific to generative \glspl{rnn} is their difficulty recovering from past errors~\cite{graves2013generating}, which \cite{krause2016multiplicative} argue arises from having hidden-state transitions that are highly correlated across possible inputs.
One approach to adapting \glspl{rnn} to have more input-dependent transition functions is to use the multiplicative "trick" \cite{sutskever2011generating}. This approximates the idea of having the input at each time synthesize a dedicated kernel of parameters dictating the transition from the previous hidden state to the next. These two approaches can be combined, as in the \gls{mlstm}~\cite{krause2016multiplicative}.

We begin by contending that, in making \glspl{rnn} multiplicative, sharing what is known as the \emph{intermediate state} does not significantly hinder performance when parameter counts are equal. 
We verify this with existing as well as new gated models on several well-known language modeling tasks.

\section{Multiplicative RNNs}
\label{sec:mrnn}

Most recurrent neural network architectures, including \gls{lstm} and \gls{gru} share the following building block:
\begin{equation}\label{eq:rnnhupdate}
    \tilde{h}_t  =  U x_t +  W h_{t-1}.
\end{equation} 
$\tilde{h}_t$ is the \emph{candidate} hidden state, computed from the previous hidden state, $h_{t-1}$, and the current input, $x_t$, weighted by the parameter matrices $W$ and $U$, respectively. 
This candidate hidden state may then be passed through gating mechanisms and non-linearities depending on the specific recurrent model.

Let us assume for simplicity that the input is a one-hot vector (one component is $1$, the rest are $0$~\cite{socher2013deep} [see p.45]), as it is often the case in \gls{nlp}. 
Then, the term $U x_t$ is reduced to a single column of $U$ and can therefore be thought of as an input-dependent bias in the hidden state transition. 
As the dependencies we wish to establish between the elements of the sequences under consideration become more distant, the term $W h_t$ will have to be significantly larger than this input-dependent bias, $U x_t$, in order to remain unchanged across time-steps.
This will mean that from one time-step to the next, the hidden-to-hidden transition will be highly correlated across possible inputs. 
This can be addressed by having more input-dependent hidden state transitions, making \glspl{rnn} more expressive.

In order to remedy the aforementioned problem, each possible input $i$ can be given its own matrix $W^{(i)}$ parameterizing the contribution of $h_t$ to $\tilde{h}_t$.
\begin{equation}\label{eq:tensorrnnh}
    \tilde{h}_t  = U x_t + \underbrace{(\sum_{i} W^{(i)}x_t^{(i)})}_{\mathbf{W}^{(x_t)}} h_{t-1}.
\end{equation} 
This is known as a \gls{trnn}~\cite{sutskever2011generating}, because all the matrices can be stacked to form a rank 3 tensor, $\mathbf{W}$. 
The input $x_t$ selects the relevant slice of the tensor in the one-hot case and a weighted sum over all slices in the dense case. 
The resulting matrix then acts as the appropriate $W$. 

However, such an approach is impractical because of the high parameter count such a tensor would entail. 
The tensor can nonetheless be approximated by factorizing it~\cite{taylor2009factored} as follows:
\begin{equation}\label{eq:fact}
    \mathbf{W}^{(x_t)} = V\textrm{diag}(W_xx_t)W_h,
\end{equation}
where $W_x$ and $W_h$ are weight matrices, and $\textrm{diag}$ is the operator turning a vector $v$ into a diagonal matrix where the elements of $v$ form the main diagonal of said matrix. 
Replacing $\mathbf{W}^{(x_t)}$ in Equation~(\ref{eq:tensorrnnh}) by this tensor factorization, we obtain
\begin{equation}\label{eq:mrnn-hhat}
\tilde{h}_t  = U x_t + V m_t,   
\end{equation} where $m_t$ is known as the \emph{intermediate state}, given by
\begin{equation}\label{eq:m}
m_t = (W_{x} x_t) * (W_{h} h_{t-1}).    
\end{equation}
Here, $*$ refers to the Hadamard or element-wise product of vectors. 
The intermediate state is the result of having the input apply a learned filter via the new parameter kernel $W$ to the factors of the hidden state. 
It should be noted that the dimensionality of $m_t$ is free and, should it become sufficiently large, the factorization becomes as expressive as the tensor. 
The ensuing model is known as a \gls{mrnn} \cite{sutskever2011generating}.

\section{Sharing intermediate states}
\label{sec:mmod}

While \glspl{mrnn} outperform simple \glspl{rnn} in character-level language modeling, they have been found wanting with respect to the popular \gls{lstm} \cite{hochreiter1997long}. 
This prompted~\cite{krause2016multiplicative} to apply the multiplicative "trick" to \gls{lstm} resulting in the \gls{mlstm}, which achieved promising results in several language modeling tasks~\cite{krause2016multiplicative}. 

\subsection{mLSTM}

Gated \glspl{rnn}, such as \gls{lstm} and \gls{gru}, use \emph{gates} to help signals move through the network. 
The value of these gates is computed in much the same way as the candidate hidden state, albeit with different parameters. 
For example, \gls{lstm} uses two different gates, $i$ and $f$ in updating its memory cell, $c_t$,
\begin{equation}\label{eq:lstm-c}
    c_t = f_t * c_{t-1} + i_t * \mathrm{tanh}(\tilde{h}_t).
\end{equation}
It uses another gate, $o$, in mapping $c_t$ to the new hidden state, $h_t$,
\begin{equation}\label{eq:lstm-h}
    h_t = o_t * \sigma (c_t),
\end{equation}
where $\sigma$ is the sigmoid function, squashing its input between 0 and 1.
$f$ and $i$ are known as forget and input gates, respectively. The forget gates allows the network to ignore components of the value of the memory cell at the past state. The input gate filters out certain components of the new hidden state. Finally, the output gates separates the memory cell from the actual hidden state. The values of these gates are computed at each time step as follows:
\begin{equation}\label{eq:lstm-i}
    i_t  = \sigma(U_i x_t + W_i h_{t-1})
\end{equation}
\begin{equation}\label{eq:lstm-f}
    f_t  = \sigma( U_f x_t + W_f h_{t-1})
\end{equation}
\begin{equation}\label{eq:lstm-o}
    o_t  = \sigma(U_o x_t + W_o h_{t-1}).
\end{equation}
Each gate has its own set of parameters to infer.
If we were to replace each $W_\star$ by a tensor factorization as in \gls{mrnn}, we would obtain a \gls{mlstm} model.
However, in the original formulation of \gls{mlstm}, there is no factorization of each would-be $\mathbf{W}_{\star}$ \emph{individually}. 
There is no separate intermediate state for each gate, as one would expect.
Instead, a single intermediate state, $m_t$, is computed to replace $h_{t-1}$ in \emph{all} equations in the system, by Eq.\ref{eq:m}. 
Furthermore, each gate has its own $V_\star$ weighting $m_t$.
Their values are computed as follows:
\begin{equation}\label{eq:mlstm-i}
    i_t  = \sigma(W_i h_{t-1} + V_i m_t)
\end{equation}
\begin{equation}\label{eq:mlstm-f}
    f_t  = \sigma(W_f h_{t-1} + V_f m_t)
\end{equation}
\begin{equation}\label{eq:mlstm-c}
    o_t  = \sigma(W_o h_{t-1} + V_o m_t).
\end{equation}
The model can therefore no longer be understood as as an approximation of the \gls{trnn}. 
Nonetheless, it has achieved empirical success in \gls{nlp}. 
We therefore try to explore the empirical merits of this shared parametrization and apply them to other \gls{rnn} architectures.

\subsection{True mLSTM}

We have presented the original \gls{mlstm} model with its shared intermediate state. 
If we wish to remain true to the original multiplicative model, however, we have to factorize every would-be $W_\star$ tensor separately. 
We have:
\begin{equation}\label{eq:tmlstm-i}
    i_t  = \sigma(U_i x_t + V_i m_{i,t})
\end{equation}
\begin{equation}\label{eq:tmlstm-f}
    f_t  = \sigma(U_f x_t + V_f m_{f,t})
\end{equation}
\begin{equation}\label{eq:tmlstm-o}
    o_t  = \sigma(U_o x_t + V_om_{o,t}),
\end{equation}
with each $m_{\star,t}$ being given by a separate set of parameters:
\begin{equation}\label{eq:tm}
m_{\star,t} = (W_{\star,x} x_t) * (W_{\star,h} h_{t-1}).    
\end{equation}
We henceforth refer to this model as \gls{tmlstm}. 
We sought to apply the same modifications to the \gls{gru} model, as \gls{lstm} and \gls{gru} are known to perform similarly~\cite{greff2016lstm,chung2014empirical,jozefowicz2015empirical}. That is, we build a \gls{tmgru} model, as well as a \gls{mgru} with a shared intermediate state.

\subsection{GRU}

The \gls{gru} was first proposed by~\cite{cho2014learning} as a lighter, simpler variant of \gls{lstm}. 
\Gls{gru} relies on two gates, called, respectively, the \emph{update} and \emph{reset} gates, and no additional memory cell. 
These gates intervene in the computation of the hidden state as follows:
\begin{equation} \label{eq:gruh}
 h_t = (1-z_t)h_{t-1}+z_t\textrm{tanh}(\tilde{h}_t),
\end{equation} 
where the candidate hidden state, $\tilde{h}_t$, is given by:
\begin{equation}\label{eq:gruhhat}
 \tilde{h}_t = U_{h}x_t + W_{h}(r_t* h_{t-1}).
\end{equation}
The update gate deletes specific components of the hidden state and replaces them with those of the candidate hidden state, thus updating its content.
On the other hand, the reset gate allows the unit to start anew, as if it were reading the first symbol of the input sequence.
They are computed much in the same way as the gates of \gls{lstm}: 
\begin{equation}\label{eq:gruz}
 z_t = \sigma(U_{z}x_t + W_{z}h_{t-1}),
\end{equation}
\vspace{-.4cm}
\begin{equation}\label{eq:grur}
 r_t = \sigma(U_{r}x_t + W_{r}h_{t-1}).
\end{equation}

\subsection{True mGRU}
 
We can now make \gls{gru} multiplicative by using the tensor factorization for $z$ and $r$:
\begin{equation}\label{eq:tmgru-z}
 z_t = \sigma(U_{z}x_t + V_{z}m_{z,t}),
\end{equation}
\begin{equation}\label{eq:tmgru-r}
 r_t = \sigma(U_{r}x_t) + V_{r}m_{r,t},
\end{equation}
with each $m_{\star,t}$ given by Eq. \ref{eq:tm}.
There is a subtlety to computing $\tilde{h}_t$, as we need to apply the reset gate to $h_{t-1}$. 
While $h_{t}$ itself is given by Eq. \ref{eq:mrnn-hhat}, $m_{h,t}$ is not computed the same way as in \gls{mlstm} and \gls{mrnn}.
Instead, it is given by:
\begin{equation}\label{eq:mgru-m}
    m_{h,t} = (W_{x} x_t) * (W_{h} (r_t* h_{t-1})).   
\end{equation}

\subsection{mGRU with shared intermediate state}

Sharing an intermediate state is not as immediate for \gls{gru}.
This is due to the application of $r_t$, which we need in computing the intermediate state that we want to share.
That is, $r_t$ and $m_t$ would both depend on each other.
We modify the role of $r_t$ to act as a filter on $m_t$, rather than a reset on individual components of $h_{t-1}$. 
Note that, when all components of $r_t$ go to zero, it amounts to having all components of $h_{t-1}$ at zero.
We have
\begin{equation}\label{eq:mgru-z}
 z_t = \sigma(U_{z}x_t + V_{z}m_{t})
\end{equation}
and
\begin{equation}\label{eq:mgru-r}
 r_t = \sigma(U_{r}x_t + V_{r}m_{t}).
\end{equation}
$\tilde{h}_t$ is given by
\begin{equation}
    \tilde{h}_t = U_{h}x_t + V_{h}(r_t * m_{t}),
\end{equation}
with $m_{t}$ the same as in \gls{mrnn} and \gls{mlstm} this time, i.e. Eq.\ref{eq:m}.
The final hidden state is computed the same way as in the original \gls{gru}~(Eq.\ref{eq:gruh}).

\section{Experiments in character-level language modeling}

\label{sec:exp}

Character-level language modeling (or character prediction) consists in predicting the next character while reading a document one character at a time. 
It is a common benchmark for \glspl{rnn} because of the heightened need for shared parametrization when compared to word-level models. 
We test \gls{mgru} on two well-known datasets, the Penn Treebank and Text8.

\subsection{Penn Treebank}
The Penn Treebank dataset~\cite{marcus1993building} comes from a series of Wall Street Journal articles written in English. 
Following~\cite{mikolov2012subword}, sections 0-20 were used for training, 21-22 for validation and  23-24 for testing, respectively, which amounts to 5.1M, 400K and 450K characters, respectively. 

The vocabulary consists of 10K lowercase words. 
All punctuation is removed and numbers were substituted for a single capital N. 
All words out of vocabulary are replaced by the token \texttt{<unk>}. \vspace{.1cm}

The training sequences were passed to the model in batches of 32 sequences. 
Following \cite{krause2016multiplicative}, we built an initial \gls{mlstm} model of 700 units.
However, we set the dimensionality of the intermediate state to that of the input in order to keep the model small.
We do the same for our \gls{mgru}, \gls{tmlstm} and \gls{tmgru}, changing only the size of the hidden state so that all four models have roughly the same parameter count.
We trained it using the Adam optimizer~\cite{adam}, selecting the best model on validation over 10 epochs. 
We apply no regularization other than a checkpoint which keeps the best model over all epochs. \vspace{.1cm}
The performance of the model is evaluated using cross entropy in \gls{bpc}, which is $log_2$ of perplexity. 

All models outperform previously reported results for \gls{mlstm}~\cite{krause2016multiplicative} despite lower parameter counts.
This is likely due to our relatively small batch size. 
However, they perform fairly similarly. 
Encouraged by these results, we built an \gls{mgru} with both hidden and intermediate state sizes set to that of the original \gls{mlstm} (700). 
This version highly surpasses the previous state of the art while still having fewer parameters than previous work. \vspace{.1cm}

For the sake of comparison, results as well as parameter counts (where available) of our models (bold) and related approaches are presented in Table \ref{tab:ptb}. \gls{mgru} and larger \gls{mgru}, our best models, achieved respectively an error of 1.07 and 0.98 \gls{bpc} on the test data, setting a new state of the art for this task.
\begin{table}[t!]
\centering
\renewcommand{\arraystretch}{1.2}
        \begin{tabular}{l|cc}
            Model & Parameter count  & $\quad$Error(BPC) \\
            \hline
            \gls{gru}~{\footnotesize\cite{empgru}} & 3M & 1.53\\
            \gls{mrnn}~{\footnotesize\cite{mikolov2012subword}} & - & 1.41 \\
            \gls{lstm}~{\footnotesize\cite{cooijmans2016recurrent}}  & - & 1.38 \\
            batch-normalized \gls{lstm}~\cite{cooijmans2016recurrent} & - & 1.32 \\
            \gls{mlstm} {\footnotesize\cite{krause2016multiplicative}}& - & 1.27 \\
            fast-slow \gls{lstm}~\cite{mujika2017fast} & 7.2M & 1.19 \\
            
            \textbf{mLSTM} & \textbf{292K} & \textbf{1.11} \\
            \textbf{tmLSTM} & \textbf{292K} & \textbf{1.09} \\
            \textbf{tmGRU} & \textbf{292K} & \textbf{1.08} \\
            \textbf{mGRU} & \textbf{292K} & \textbf{1.07} \\
            \textbf{larger mGRU} & \textbf{2.1M} &\textbf{0.98}\\
        \end{tabular}
    \caption{Test set error on Penn Treebank and parameter counts in character-level language modeling}
    \label{tab:ptb}
\end{table}

\subsection{Text8}
The Text8 corpus~\cite{hutter2006human} comprises the first 100M plain text characters in English from Wikipedia in 2006. 
As such, the alphabet consists of the 26 letters of the English alphabet as well as the space character. 
No vocabulary restrictions were put in place. 
As per~\cite{mikolov2012subword}, the first 90M and 5M characters were used for training and validation, respectively, with the last 5M used for testing. \vspace{.1cm}

\begin{table}[t!]
\centering
\renewcommand{\arraystretch}{1.2}
        \begin{tabular}{l|cc}
            Model & Parameter count & $\quad$ Error (BPC) \\
            \hline
            \gls{gru}~{\footnotesize\cite{empgru}} & 5M & 1.53 \\
            \gls{mrnn}~{\footnotesize\cite{mikolov2012subword}} & -  & 1.54 \\
            \gls{lstm}~{\footnotesize\cite{cooijmans2016recurrent}}  & - & 1.43 \\
            \gls{mlstm}~{\footnotesize\cite{krause2016multiplicative}}  & 20M & 1.42\\
            \textbf{mLSTM} & \textbf{133K} & \textbf{1.37} \\
            batch-normalized \gls{lstm}\cite{cooijmans2016recurrent}  & - & 1.36 \\
            \textbf{tmGRU} & \textbf{133K} & \textbf{1.35} \\
            \textbf{tmLSTM} & \textbf{133K} & \textbf{1.35} \\
            \textbf{mGRU} & \textbf{133K} & \textbf{1.35} \\
            large \gls{mlstm} \cite{krause2016multiplicative} & 46M & 1.27\\
            \textbf{larger \gls{mgru}} & \textbf{877K} & \textbf{1.21} \\
            \gls{lstm}~{\footnotesize\cite{krause2017dynamic}*}  & 45M  & 1.19 \\
        \end{tabular}
        
    \caption{Test set error on Text8 and parameter counts in character-level language modeling}
   \label{tab:text8}
\end{table}

Encouraged by our results on the Penn Treebank dataset, we opted to use similar configurations. 
However, as the data is one long sequence of characters, we divide it into sequences of 200 characters. 
We pass these sequences to the model in slightly larger batches of 50 to speed up computation. 
Again, the dimensionality of the hidden state for \gls{mlstm} is set at 450 after the original model, and that of the intermediate state is set to the size of the alphabet.
The size of the hidden state is adjusted for the other three models as it was for the PTB experiments.
The model is also trained using the Adam optimizer over 10 epochs.

The best model as per validation data over 10 epochs achieves 1.40 \gls{bpc} on the test data, slightly surpassing an \gls{mlstm} of smaller hidden-state dimensionality (450) but larger parameter count.
Our results are more modest, as are those of the original \gls{mlstm}.
Once again, results do not vary greatly between models.

As with the Penn Treebank, we proceed with building an \gls{mgru} with both hidden and intermediate state sizes set to 450.
This improves performance to 1.21 \gls{bpc}, setting a new state of the art for this task and surpassing a large \gls{mlstm} of 1900 units from \cite{krause2016multiplicative} despite having far fewer parameters (45M to 5M).

For the sake of comparison, results as well as parameter counts of our models and related approaches are presented in Table \ref{tab:text8}. It should be noted that some of these models employ \emph{dynamic evaluation}~\cite{graves2013generating}, which fits the model further during evaluation. We refer the reader to \cite{krause2017dynamic}. These models are indicated by a star.

\section{Conclusion}
\label{sec:conc}

We have found that competitive results can be achieved with \glspl{mrnn} using small models.
We have not found significant differences in the approaches presented, despite added non-intuitive parameter-sharing constraints when controlling for model size. 
Our results are restricted to character-level language modeling.
Along this line of thought, previous work on \glspl{mrnn} demonstrated their increased potential when compared to their regular variants \cite{sutskever2011generating,krause2016multiplicative,radford2017learning}. 
We therefore offer other variants as well as a first investigation into their differences.
We hope to have evinced the impact of increased flexibility in hidden-state transitions on \glspl{rnn} sequence-modeling capabilities.
Further work in this area is required to transpose these findings into applied tasks in \gls{nlp}.

\bibliography{ref}
\bibliographystyle{splncs04}

\appendix

\end{document}